\documentclass[letterpaper, 10 pt, conference]{ieeeconf}  
\IEEEoverridecommandlockouts                              
\makeatletter
\let\NAT@parse\undefined
\makeatother

\usepackage{graphics} 
\usepackage{epsfig} 
\usepackage{amsmath} 
\usepackage{booktabs}

\usepackage{algorithm}
\usepackage{algpseudocode} 

\usepackage{hyperref}

\title{\LARGE \bf
The Robot's Inner Critic: Self-Refinement of Social Behaviors \\ through VLM-based Replanning
}

\author{Jiyu Lim\textsuperscript{1}, Youngwoo Yoon\textsuperscript{2$\dagger$}, and Kwanghyun Park\textsuperscript{1$\dagger$} 
\thanks{$^{1}$\raggedright KwangWoon University, Republic of Korea. {\tt\small jsjydk25343@gmail.com, akaii@kw.ac.kr}}%
\thanks{\textsuperscript{2}ETRI, Republic of Korea. {\tt\small youngwoo@etri.re.kr}}
\thanks{\textsuperscript{$\dagger$}Corresponding authors}
\thanks{This work was supported by Korea Institute for Advancement of Technology(KIAT) grant funded by the Korea Government(MOTIE)(RS-2024-00406796, HRD Program for Industrial Innovation); by Industrial Strategic Technology Development Program funded by MOTIE (No. 20023495); by the National Research Council of Science \& Technology(NST) grant by the Korea government(MSIT) (No. GTL25041-000).}%
}

\begin{document}

\maketitle
\thispagestyle{empty}
\pagestyle{empty}

\begin{abstract}

Conventional robot social behavior generation has been limited in flexibility and autonomy, relying on predefined motions or human feedback. This study proposes CRISP (Critique-and-Replan for Interactive Social Presence), an autonomous framework where a robot critiques and replans its own actions by leveraging a Vision-Language Model (VLM) as a `human-like social critic.' CRISP integrates (1) extraction of movable joints and constraints by analyzing the robot's description file (e.g., MJCF), (2) generation of step-by-step behavior plans based on situational context, (3) generation of low-level joint control code by referencing visual information (joint range-of-motion visualizations), (4) VLM-based evaluation of social appropriateness and naturalness, including pinpointing erroneous steps, and (5) iterative refinement of behaviors through reward-based search. This approach is not tied to a specific robot API; it can generate subtly different, human-like motions on various platforms using only the robot's structure file. In a user study involving five different robot types and 20 scenarios, including mobile manipulators and humanoids, our proposed method achieved significantly higher preference and situational appropriateness ratings compared to previous methods. This research presents a general framework that minimizes human intervention while expanding the robot's autonomous interaction capabilities and cross-platform applicability. Detailed result videos and supplementary information regarding this work are available at \url{https://limjiyu99.github.io/inner-critic/}.

\end{abstract}

\section{Introduction}
For robots to integrate naturally into human daily life, the ability to generate appropriate behaviors for human interaction is essential \cite{mohammadi2019designing,porfirio2020transforming}. For instance, if a person is hesitating in front of a door with their hands full, a socially aware robot should recognize the situation and open the door. Similarly, if a person approaches and greets the robot cheerfully, it is natural for the robot to greet them back. Conventional approaches to generating social behaviors in robots have often been rigid, relying on rule-based or template-based methods \cite{aly2013model,david2022interaction}. These methods struggle to adapt to diverse situations and are limited to repeating predefined motions.
Recent studies have moved away from these rigid approaches, instead using Large Language Models (LLMs) to generate combinations of predefined action functions, aiming for more flexible responses in various interaction scenarios \cite{mahadevan2024generative,park2024towards}. However, because LLM-generated responses are not always consistent for a given situation, their guidance doesn't guarantee a successful outcome every time. Therefore, research on correcting the erroneous behaviors of LLMs has relied on methods involving direct human observation and intervention \cite{weber2018shape,mahadevan2024generative}. Yet, receiving direct human feedback during an interaction can diminish the user's satisfaction, and collecting and learning from human feedback data is a costly and time-consuming process \cite{tian2024maximizing}. Relying on human feedback to correct robot behavior fundamentally limits the robot's autonomy.

\begin{figure}[t]
    \centering
    \includegraphics[width=9cm]{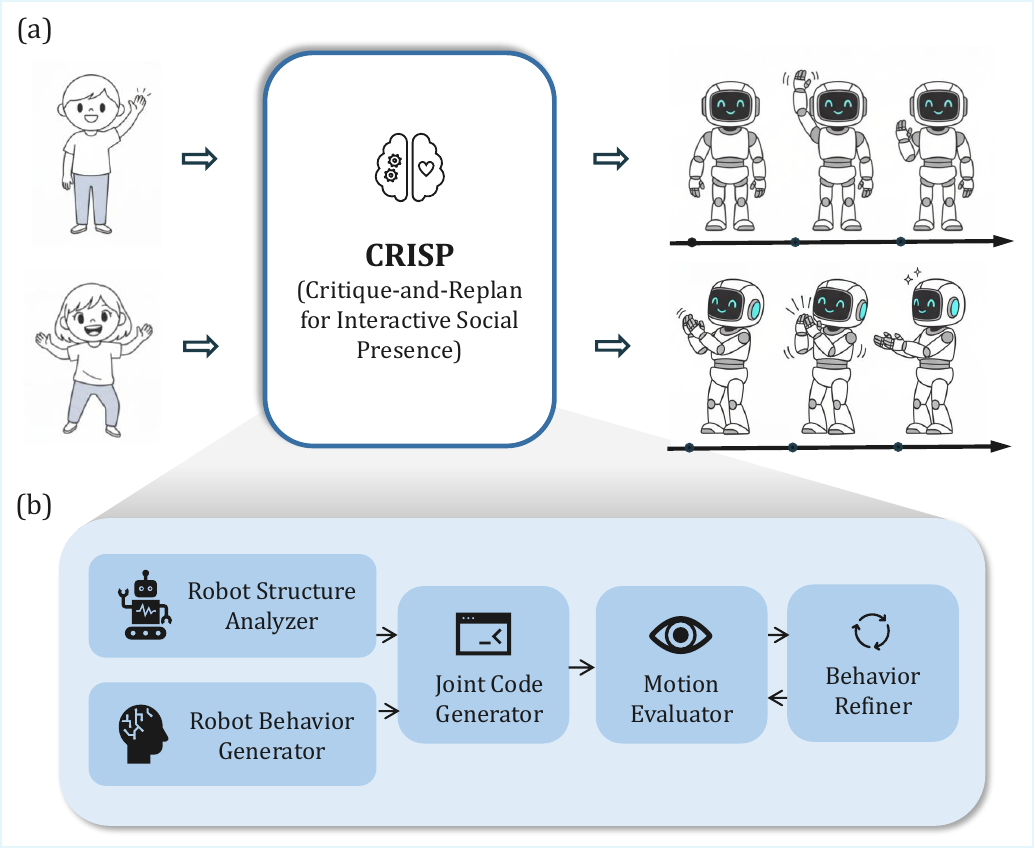}
    \vspace{-8mm}
    \caption{Overview of the proposed framework. (a) The robot generates socially appropriate responses to human actions. For example, if a person waves to greet, the robot waves back. If a person dances joyfully, the robot generates an action of clapping and cheering. (b) Given a robot's structural file, the system produces low-level joint control code, which is refined through iterative VLM evaluation for natural and context-appropriate behavior.}
    \label{fig:intro}
    \vspace{-5mm}
\end{figure}

To address this, we propose a framework that enables a robot to autonomously replan its social behaviors without human intervention by using a VLM (Fig.~\ref{fig:intro}). In the field of task planning with LLMs, research on self-correction and replanning for failed actions is actively being pursued \cite{mei2024replanvlm,chen2025robogpt}. Inspired by this, we utilize a VLM as a `human-like social critic,' creating a cyclic process where the robot observes its own behavior, identifies awkwardness, and refines its actions for the better. This autonomous cycle of `generate-evaluate-regenerate' maximizes the robot's autonomous reasoning capabilities, minimizing human intervention and allowing it to independently enhance its interaction skills. 

Furthermore, existing studies that use predefined robot primitives (e.g., wave hand, nod head) \cite{mahadevan2024generative,park2024towards} are often tied to a specific robot, requiring developers to implement motions for each new platform. This also leads to mechanical and repetitive behaviors in similar situations. In contrast, humans perform the same action in subtly different ways each time. For example, the act of waving a hand is never executed with the exact same angle or speed. For robots to achieve better interaction with humans, it is crucial to emulate this flexibility \cite{destephe2015walking}, as repetitive and predictable motions often diminish long-term engagement \cite{suguitan2020moveae}.

This study achieves greater interaction flexibility by constructing a framework capable of low-level control (generating time-series joint values) rather than high-level control of predefined robot actions. High-level control significantly lacks flexibility, as it cannot execute novel behaviors that have not been predefined \cite{shentu2024llms}. To overcome this, we adopt a low-level control approach. Existing approaches to low-level robot control have largely relied on learning from extensive datasets \cite{williams2017information,kim2024openvla}. However, datasets for social interactions are difficult to obtain, and the learning process is immensely costly. Our research reduces the need for task-specific data collection and training, enabling low-level control through few-shot prompting \cite{dong2024survey}. Moreover, the system is designed to be robot-agnostic; given a robot's structure file (e.g., MJCF), an LLM can analyze its joint information and generate corresponding movements.

Our system comprises five core components. First, the LLM analyzes the structural file of the target robot (we used MJCF for the MuJoCo simulator) to gather information about its structure, limitations, and movable joints. This joint information is then converted into images that represent the spatial awareness associated with joint movements. Second, it generates appropriate robot behaviors within the context of human-robot interaction. It breaks down a socially appropriate response to a human action into several steps, creating motions tailored to each robot using the joint information and movement limits obtained in the first step. Third, it takes the joint information and its visual representation from the first step and the action steps from the second step as input, and converts them into appropriate joint movement code. Fourth, a VLM evaluates the social appropriateness of the entire motion code and autonomously proposes refinements. Fifth, the VLM observes the actual execution of this code and refines it through multiple replanning iterations. By providing only the current situation and the target robot file as input, our system generates a suitable behavior as output. We assume that a description of the situational context is given, and our method is an offline generation process involving iterative simulation and VLM evaluation.

The main contributions of this study can be summarized as follows:
\subsubsection{Autonomous Behavior Evaluation and Refinement using a VLM} 
We propose an autonomous `generate–evaluate–regenerate' cycle in which a VLM functions as a `human-like social critic.' This enables robots to self-assess and refine the social appropriateness of their actions without explicit human feedback, enhancing autonomy in social interaction.
\subsubsection{Achieving Flexibility through Low-level Control} 
We move beyond predefined action APIs by implementing low-level control directly from a robot's structural file (e.g., MJCF). This design allows the system to generate flexible, natural, and subtly varied human-like motions that are not tied to any specific platform, ensuring broad adaptability across robots.
\subsubsection{Comprehensive Evaluation across Platforms and Scenarios} 
We validated our system on five robot platforms, from mobile manipulators to humanoids, across 20 interaction scenarios. User studies confirmed significant improvements in preference and situational appropriateness over prior methods, and an ablation study demonstrated the contribution of each system component.

\section{Related Work}

\subsection{Approaches for Robot Social Behavior Generation}
Research on robot social behavior generation can be broadly classified into rule-based, template-based, and data-driven approaches \cite{oralbayeva2024data}. 
Rule-based methods determine behavior through explicitly defined rules or logical operations \cite{aly2013model,huang2012robot}, while template-based methods learn generalized patterns from recorded interactions and reuse them as templates \cite{david2022interaction}. These approaches struggle to produce novel behaviors and have limited expressiveness. 
Data-driven methods address this limitation by learning interaction logic from accumulated datasets \cite{yoon2019robots}, but they face poor generalization when applied to new contexts or robot platforms.

Recent studies explore using Large Language Models (LLMs) to generate behaviors \cite{mahadevan2024generative,park2024towards}. While promising, most rely on predefined functions or action primitives, restricting the flexibility of generated motions. For instance, SAMALM \cite{wang2025multi} produced low-level signals with an LLM but was limited to navigation. 
In contrast, our study extends LLM-based methods by introducing a VLM-based replanning framework that leverages robot model descriptions to generate situationally appropriate social behaviors via low-level control across multiple platforms.

\begin{figure*}[t!]
  \centering
  \includegraphics[width=\textwidth]{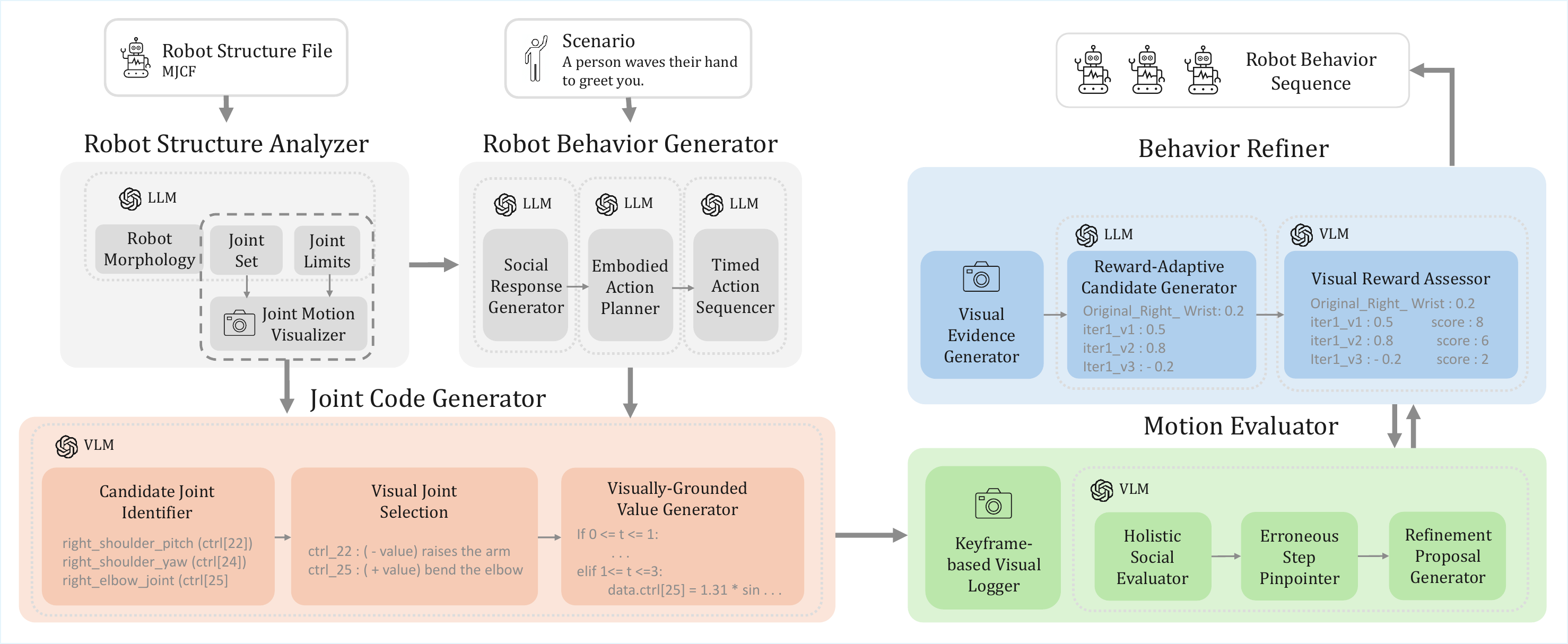}
  \vspace{-8mm}
  \caption{Overview of the proposed social behavior generation framework. The framework parses robot morphology, plans social behaviors, generates low-level joint commands, and uses a VLM critic to iteratively evaluate and replan actions.}
  \label{fig:overview}
  \vspace{-2mm}
\end{figure*}

\subsection{Robot Behavior Evaluation and Autonomous Refinement}
Replanning methods using VLMs, where robots assess visual outcomes and adjust actions, have been studied in task-oriented domains with clear success criteria \cite{mei2024replanvlm,skreta2024replan}. 
However, in domains requiring subjective judgment, such as social behavior, evaluation and refinement still rely heavily on human feedback \cite{mahadevan2024generative,huang2025emotion}. 
Collecting such feedback is time-consuming, costly, and requires large-scale preference data \cite{tian2024maximizing}. 
Research that enables autonomy in replanning subjective social behaviors remains limited. 
Our work addresses this gap by proposing a framework in which a VLM evaluates the social appropriateness of behaviors with human-like judgment and autonomously refines them.

\section{Proposed Method}
In this work, we propose CRISP (Critique-and-Replan for Interactive Social Presence), a framework for autonomously generating and refining flexible social behaviors that are not dependent on a specific robot. Fig. \ref{fig:overview} illustrates the overall system architecture. The system consists of five core modules: a Robot Structure Analyzer, a Robot Behavior Generator, a Joint Code Generator, a Motion Evaluator, and a Behavior Refiner. In this section, we will describe the technical details of each module in sequence.

\subsection{Robot Structure Analyzer}
The Robot Structure Analyzer is responsible for automatically analyzing the robot's physical structure and kinematic constraints, and extracting key information for subsequent modules. It is designed to automate the manual analysis and motion function design process for each robot, which was required in previous studies, thereby ensuring the system's general applicability.
The analysis process consists of two stages.

\textbf{Structural Specification Analysis using LLM.} First, the specification file containing the robot's structural information (in this study, MJCF, as we use the MuJoCo simulator) is taken as input. The LLM parses this file to automatically extract the following three pieces of information:
\begin{itemize}
\item Joint Set ($J$): It identifies the set of all actively controllable joint axes of the robot, $J = \{j_1, j_2, ..., j_n\}$.
\item Joint Limits: For each joint $j_i\in J$, it extracts the minimum and maximum permissible angle range $[L_i^{\min}, L_i^{\max}]$.
\item Robot Morphology: It provides a textual summary of the robot's overall degrees of freedom and the connectivity of its body parts. For example, for the G1 robot shown in Fig. \ref{fig:robots}, it generates information like ``Each arm has 7-DOF, and the waist can move along the pitch, roll, and yaw axes.'' This morphological information is later used by the Robot Behavior Generator to devise actions appropriate for that specific robot.
\end{itemize} 

\begin{figure}[!b]
  \vspace{-4mm}
  \centering
  \includegraphics[width=8.5cm]{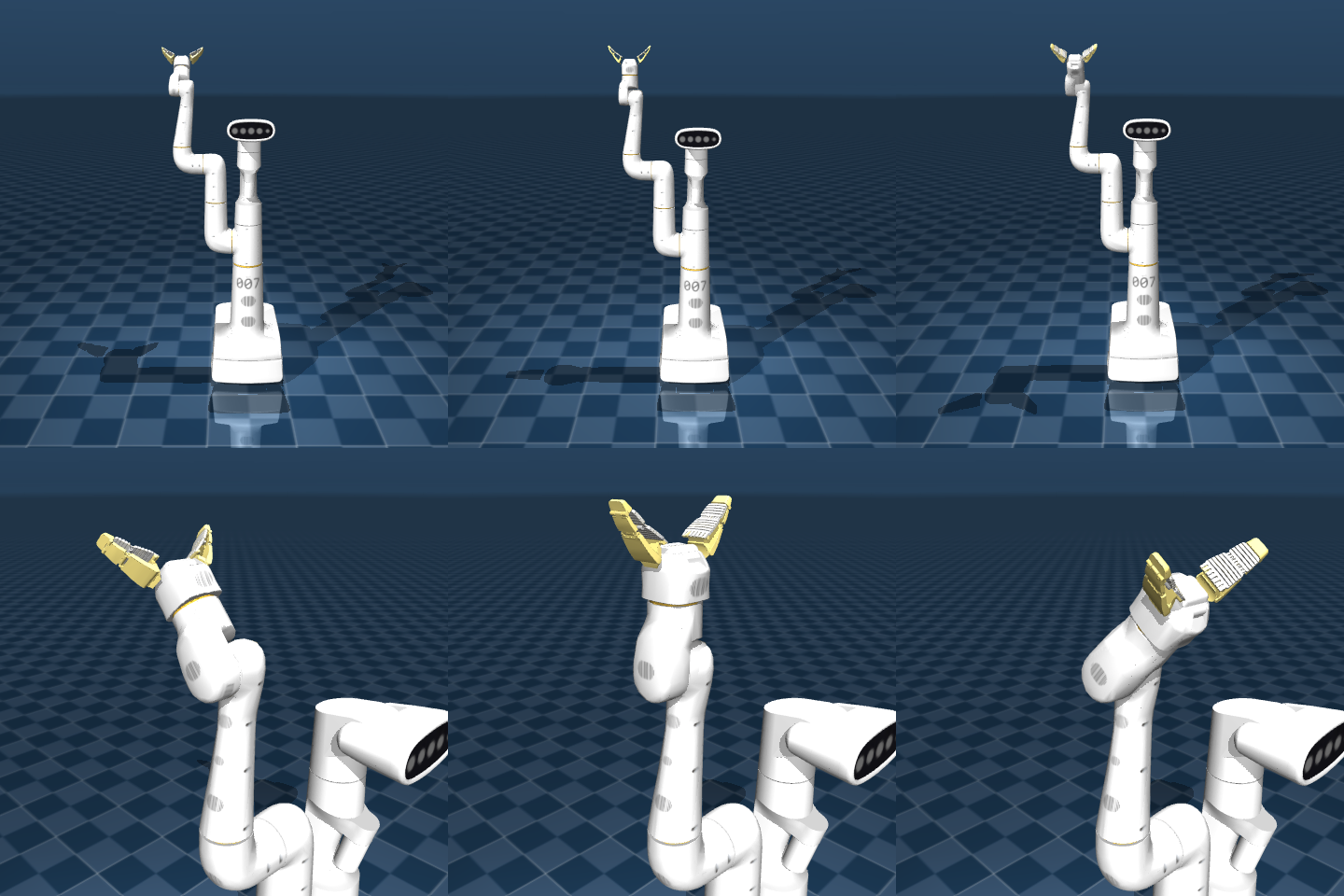}
  \vspace{-2mm}
  \caption{Visualization of robot joint range of motion. Full-body and zoomed-in images for positive, zero, and negative values of the Everyday robot's wrist joint.}
  \label{fig:joint_viz}
\end{figure}

\textbf{Simulation-based Visualization of Joint Movements.}
The extracted joint information ($J$ and $[L_i^{\min}, L_i^{\max}]$) is used to build a visual dataset $D_{\text{visual}}$. This is to overcome the limitation that LLMs struggle to predict intuitive spatial relationships or the results of movements from text information alone. Fig. \ref{fig:joint_viz} shows an example for a wrist joint. For each joint $j_i$, we capture three configurations: the simulator default (center panel) and two samples at $\pm 50\%$ of the joint’s half-range around the default value (side panels). 
Each panel includes both a full-body view and a zoomed‑in crop of the actuated region to reveal subtle local motion. 
The resulting set $D_{\text{visual}}=\{I_1,\dots,I_n\}$, where each $I_i$ is the set of visualization images for joint $j_i$, serves as a visual dictionary that maps joint-value changes to spatial movements, which the VLM in the Joint Code Generator consults to produce spatially plausible code.

\subsection{Robot Behavior Generator}
This module translates a human social action in a given context into a concrete sequence of robot behaviors. Inspired by the work of Mahadevan et al. \cite{mahadevan2024generative}, we adopted the idea of having an LLM generate actions while considering the robot's physical constraints. We further enhanced this by adding the capability to plan the estimated execution time for each action step, thereby improving the realism and flexibility of the generated behavior. First, the current interaction context (e.g., ``A person waves their hand to greet you.'') and the robot's morphology and capabilities, provided by the Robot Structure Analyzer, are input to the LLM. Based on this, the LLM translates the human social action (e.g., ``Smile warmly and wave back to acknowledge their greeting.'') into a form that the robot can actually perform. For example, for a robot incapable of smiling or turning its head, the action is specified as ``Turn the torso to face the person, raise one arm, and use the wrist joint to wave.'' This translated robot action is then fed back into the LLM, which breaks down the entire behavior into smaller steps and assigns an estimated execution time to each. This results in a plan with temporal information, such as ``Step 1: Rotate the waist slightly to the left to create a welcoming posture (0-1 s),'' ensuring a natural flow of motion. Thus, this module generates a flexible social behavior plan that considers the robot's characteristics and temporal flow, and passes it to the next module.

\subsection{Joint Code Generator}

This module converts the step-by-step behavior plan (denoted as $S_k$, where $k$ is the index of each behavior step) from the Robot Behavior Generator into executable code driving joint movement. It uses the set of movable joints ($J$), their limits ($[L_i^{\min}, L_i^{\max}]$), and the visual dataset ($D_{\text{visual}}$) from the Robot Structure Analyzer as input. The module employs a Vision-Language Model (VLM) that uses Chain-of-Thought (CoT) reasoning \cite{wei2022chain} to determine the appropriate joint position value $v^*$ for each behavior step $S_k$.

\textbf{Identify Relevant Joints.} The VLM first interprets the behavior description $S_k$ (e.g., ``Tilt the head upward.'') to identify a set of candidate joints $J_{candidate}\subseteq J$ relevant to the action. For this example, `head yaw' and `head pitch' would be selected as candidates.

\textbf{Joint Selection through Visual Analysis.} The VLM then consults the visual data $\{I_m | j_m \in J_{candidate}\}$ for the candidate joints. By visually analyzing the movements depicted in these images, it selects the joint $j^*$ that best matches the description $S_k$. For instance, to ``Tilt the head upward,'' the VLM chooses the `head pitch' joint, whose image ($I_{HeadPitch}$) shows vertical motion, over the `head yaw' joint ($I_{HeadYaw}$), which shows horizontal motion.

\textbf{Value Generation based on Visual Reference.} Using the visual data $I_{j^*}$ for the selected joint, the VLM infers the final value $v^*$ that matches the required intensity and direction from $S_k$. It references the positive and negative movement examples within $I_{j^*}$. For example, if the positive sample in $I_{HeadPitch}$ shows a moderate tilt (an angular position of 0.5 radians), but the command requires tilting further, the VLM will generate a value greater than 0.5. All generated values $v^*$ are constrained within the joint's limits, $[L_{j^*}^{\min}, L_{j^*}^{\max}]$.

This process is repeated for all behavior steps, generating a set of joint control commands $C_k = \{(j^*, v^*), \dots\}$ for each step. Finally, the complete control sequence for the entire behavior, $C = \{C_1, C_2, \dots, C_m\}$, is passed to the next module.

\begin{figure}[!t]
  \centering
  \includegraphics[width=8.5cm]{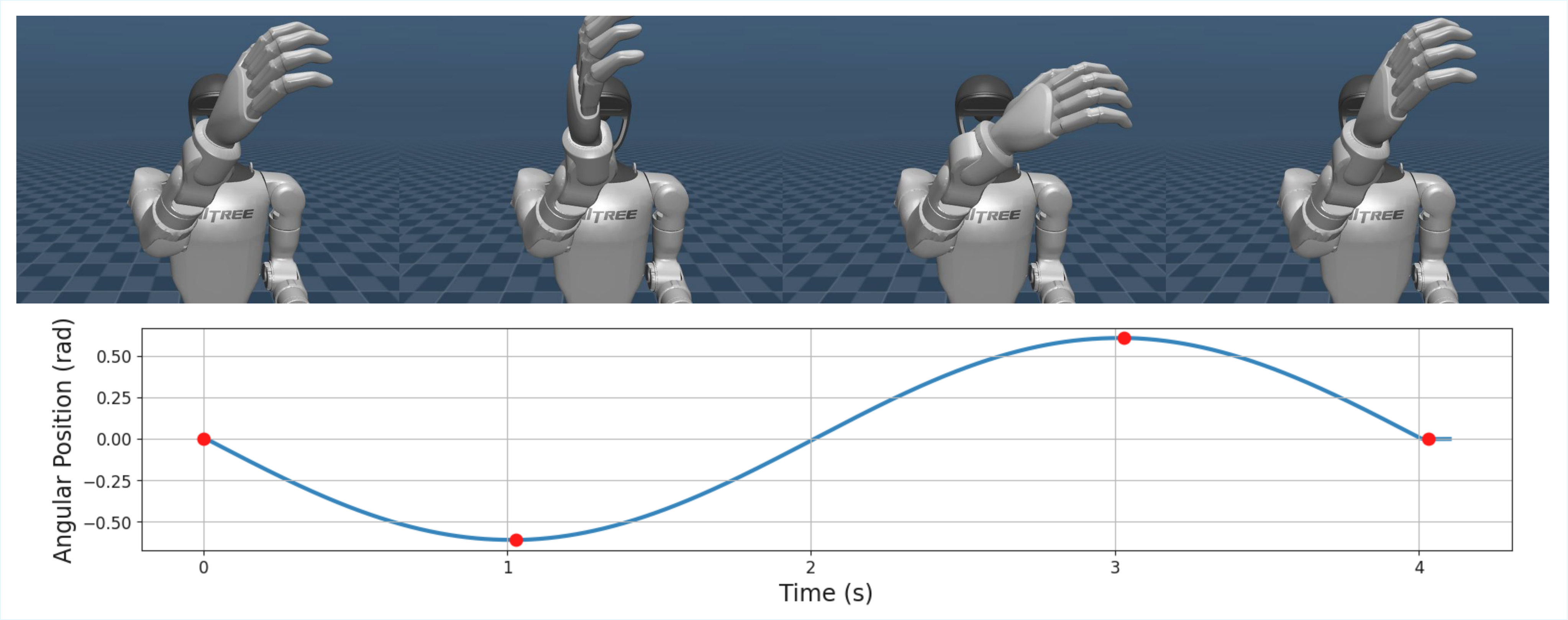}
  \vspace{-2mm}
  \caption{Example of keyframe capture for continuous movement. Keyframes are captured where the angular velocity is zero. The graph below shows the change in wrist angle over time, with red dots indicating the captured moments.}
  \label{fig:keyframe}
  \vspace{-4mm}
\end{figure}

\subsection{Motion Evaluator}

This module evaluates the social appropriateness of the generated joint code and proposes refinements if necessary. The process involves two main stages: visual log generation and a three-step evaluation.

\textbf{Visual Log Generation.} First, the generated joint code is executed in a simulator to create a visual log $V$, composed of keyframe images for each behavior step ($S_k$). Keyframes are captured as follows:
\begin{itemize}
    \item For a single target pose (e.g., moving a shoulder to 0.7), one keyframe is captured upon completion.
    \item For continuous motion (e.g., a wave), keyframes are captured at points of zero angular velocity to represent the motion's flow. (See Fig. \ref{fig:keyframe})
\end{itemize}
Each keyframe includes both full and zoomed-in views, similar to the Robot Structure Analyzer. This visual log $V$ is then passed to the VLM for evaluation.

\textbf{Holistic Social Appropriateness Evaluation.} First, the VLM performs a holistic evaluation using only the context and visual log ($V$). It assesses the motion based on social appropriateness, human-likeness, and gesture completeness as guided in the prompt (please refer to the supplementary webpage for the full prompt). For instance, if a `hand wave' lacks wrist motion, the VLM might critique it for not being human-like: ``The wrist needs to be waved for a natural greeting.'' If no issues are found, the behavior is deemed ``highly appropriate,'' and the process concludes.

\textbf{Pinpointing the Erroneous Step.} If a critique is generated, the VLM then pinpoints the erroneous step. It semantically compares the critique with each step in the behavior plan $\{S_1, \dots, S_m\}$ to find the most relevant step index, $k^*$. If multiple steps are relevant, the earliest one is chosen to efficiently manage cascading effects on subsequent actions. We empirically validate this design choice by comparing against a variant that refines the entire plan at once in Section \ref{ssec:ab}.

\textbf{Generating Specific Refinement Proposals.} Finally, the VLM generates a specific refinement plan, $A_{refine}$, for the problematic step $S_{k^*}$ based on the critique. It first identifies all related joints as candidates. By referencing their visual data ($D_{\text{visual}}$), it determines the best way to modify the motion. The resulting plan, $A_{refine}$, takes one of three forms:
\begin{itemize}
    \item \textit{Adjust}: Modify the value of an existing joint command.
    \item \textit{Delete}: Remove an unnecessary joint command.
    \item \textit{Add}: Add a new, missing joint command.
\end{itemize}
This refinement plan is then recorded in the Behavior Artifact file, making it available to the next module.

\subsection{Behavior Refiner}

This module explores the optimal robot behavior by modifying the joint code based on refinement proposals ($A_{refine}$) from the Motion Evaluator and iteratively evaluating the results. This process operates similarly to concepts in Reinforcement Learning, and we formalize it as the Reward-based Adaptive Search (RAS) algorithm. We define each iteration of the refinement process as $t$.

\textbf{Initialization and Candidate Value Generation (Initialization: $t=0$).} When the Behavior Artifact file and the refinement instruction ($A_{refine}$) are received, the LLM generates an initial set of candidates $V_0 = \{v_{0,1}, v_{0,2}, v_{0,3}\}$ for the erroneous step ($S_k*$) where the subscript $0$ denotes the initialization iteration ($t=0$). In the first iteration, values are explored with a wide range ($\sigma_0$) within the joint's limits $[L_{j^*}^{\min}, L_{j^*}^{\max}]$. Two of the three candidates follow the direction from $A_{refine}$, while the third is generated in the opposite direction to avoid local optima. Each value is saved to the Step Log file for that step.

\textbf{Evaluation and Iterative Exploration ($t \ge 1$).}
\subsubsection{Visual Evidence Generation and Reward Evaluation}
The generated candidates are passed to the capture module to create a visual log. For fluid motions, keyframes are captured when the angular velocity is zero. Each image is fed into the VLM, which assesses the movement by comparing it to the given goal and assigns a reward score $R(v_{t-1, i})$ between 1 and 10.
\begin{itemize}
    \item $R_i \in [8, 10]$: Successful movements.
    \item $R_i \in [5, 7]$: Correct direction but incomplete.
    \item $R_i \in [3, 4]$: Incorrect but not opposite direction.
    \item $R_i \in [1, 2]$: Opposite direction.
\end{itemize}
The value with the highest reward and the reward itself are defined as $v_{t-1}^*$ and $R_{t-1}^*$, respectively:
$$R_{t-1}^* = \max_{i} R(v_{t-1, i}),$$
$$v_{t-1}^* = \arg\max_{v \in V_{t-1}} R(v).$$

\subsubsection{Termination Condition Check \& Reward-based Search Width Update}
The algorithm terminates if $R_{t-1}^* \ge \tau$ (we set $\tau$ to 8 in our experiment), at which point $v_{t-1}^*$ is finalized in the Behavior Artifact file and returned to the Motion Evaluator.

If the reward is less than $\tau$, the VLM adaptively adjusts the search width $\sigma_t$ based on the maximum reward $R_{t-1}^*$.
$$
\sigma_t =
\begin{cases}
\alpha \cdot \sigma_{\text{base}} & \text{if } 5 \le R_{t-1}^* < \tau \quad \text{(fine-grained search)} \\
\beta \cdot \sigma_{\text{base}} & \text{if } R_{t-1}^* < 5 \quad \text{(broad search)}
\end{cases}
$$
Here, $\sigma_{\text{base}}$ is the base search width, and $\alpha$ and $\beta$ are scaling factors for fine-grained and broad searches (e.g., $0<\alpha<1, \beta>1$). For instance, if a reward of 5 or higher is received, the search width is reduced for fine-tuning. Conversely, if the reward is low, the width is increased for a broader search. The parameters $\alpha_{base}$, $\alpha$ and $\beta$ are empirically determined for each robot, considering the general characteristics of its joints' ranges of motion. for G1 robot whose key joints operate in a wide range ($\approx$4 rad on average), we set $\sigma_{base}$=0.6, $\alpha$=0.4, and $\beta$=1.5.

\subsubsection{New Candidate Value Generation and Iteration}
In each iteration, we sample three new candidates---a number chosen to balance exploration diversity against evaluation cost---from a normal distribution $\mathcal{N}(v_{t-1}^*, \sigma_t^2)$ and clipped to be within the joint's limits [$L^{\min}, L^{\max}$].
$$\tilde{v}_{t,i} \sim \mathcal{N}(v_{t-1}^*, \sigma_t^2) \quad \text{for } i=1,2,3$$
$$v_{t,i} = \text{clip}(\tilde{v}_{t,i}, L^{\min}, L^{\max})$$
This new candidate set $V_t$ is then sent back for repeated evaluation.

\subsubsection{Search Failure Diagnosis}
If rewards are consistently low ($R_{t-1}^* \leq 2$), the process returns to the Motion Evaluator to search for a new joint. All refinement processes and reward histories are recorded in the Behavior Artifact file to manage the search history and prevent infinite loops.

\begin{figure}[!t]
  \centering
  \includegraphics[width=8.5cm]{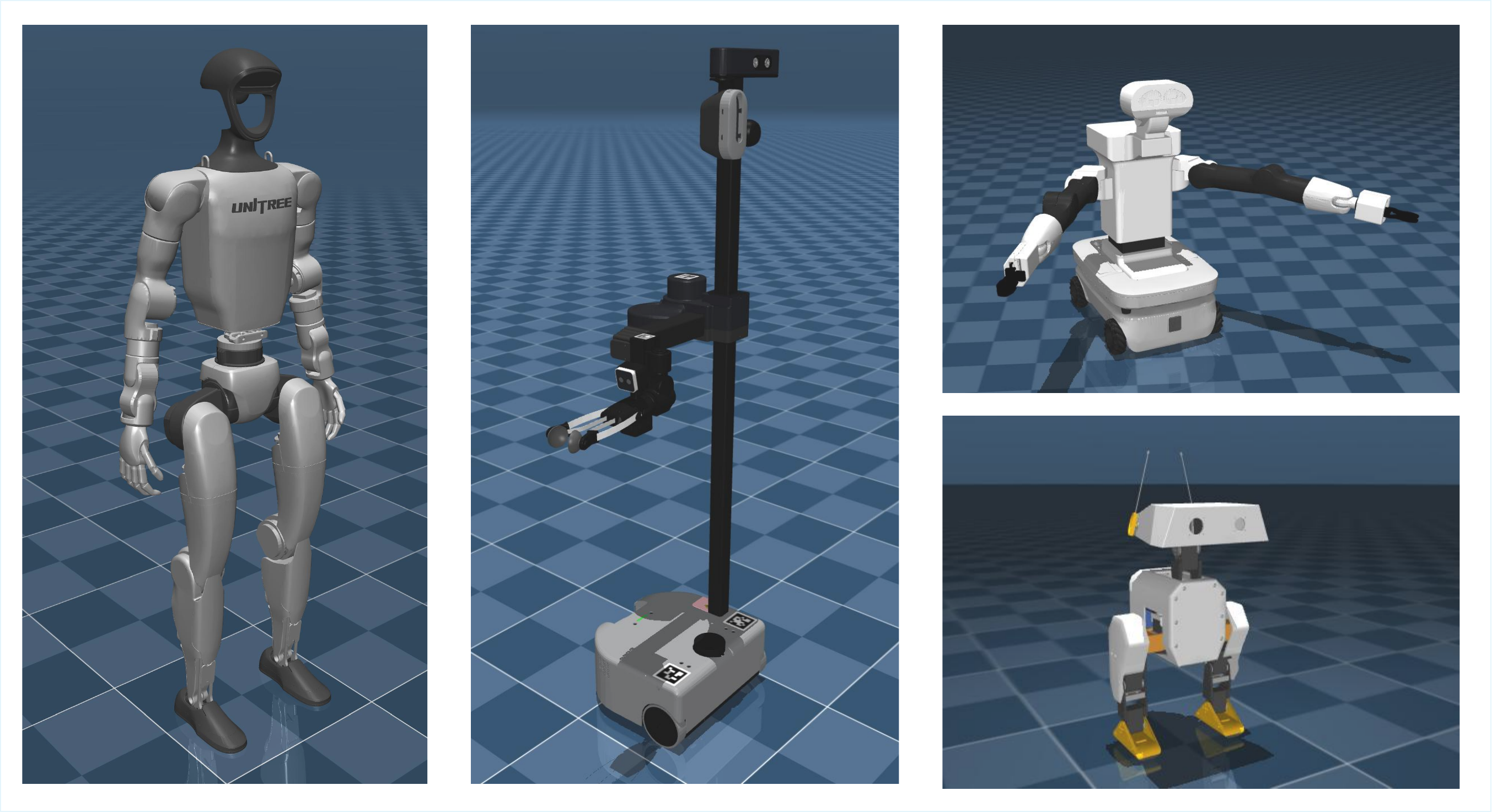}
  \vspace{-3mm}
  \caption{Types of robots used in the experiment: Unitree G1, Stretch 3 (left), TIAGo (top right), Open Mini Duck (bottom right). The Everyday robot is shown in Fig. \ref{fig:joint_viz}. The proposed method can generate social behaviors for a wide range of robots, from simple ones to humanoids.}
  \label{fig:robots}
  \vspace{-4mm}
\end{figure}
\section{Experiments}

To verify whether our proposed VLM-based replanning system, CRISP, provides a superior user experience, we conducted a human-subject evaluation. For the experiments in this paper, we used OpenAI's GPT-4o (Accessed in August 2025) as both the VLM and LLM, with the temperature set to 0.

\subsection{Experiment Setup}
The user evaluation was conducted by having participants watch video clips of robot behaviors generated by different systems for specific scenarios and then rate their quality. To assess the generalization performance of the proposed system across robots with different morphologies, we used the following five types of robots: Google Research's Everyday Robot (mobile manipulator), Hello Robot Stretch 3 (mobile manipulator), Open Mini Duck (an open-source project based on Disney Research's BDX Droid, a simple duck-shaped robot), PAL Robotics TIAGo (dual-arm manipulator), and Unitree G1 (humanoid robot) (see Fig. \ref{fig:robots} and \ref{fig:joint_viz}).

\begin{figure}[t!]
  \centering
  {\small \textbf{Scenario:} ``A person waves their hand to greet you.''} \\
  \includegraphics[width=8.5cm]{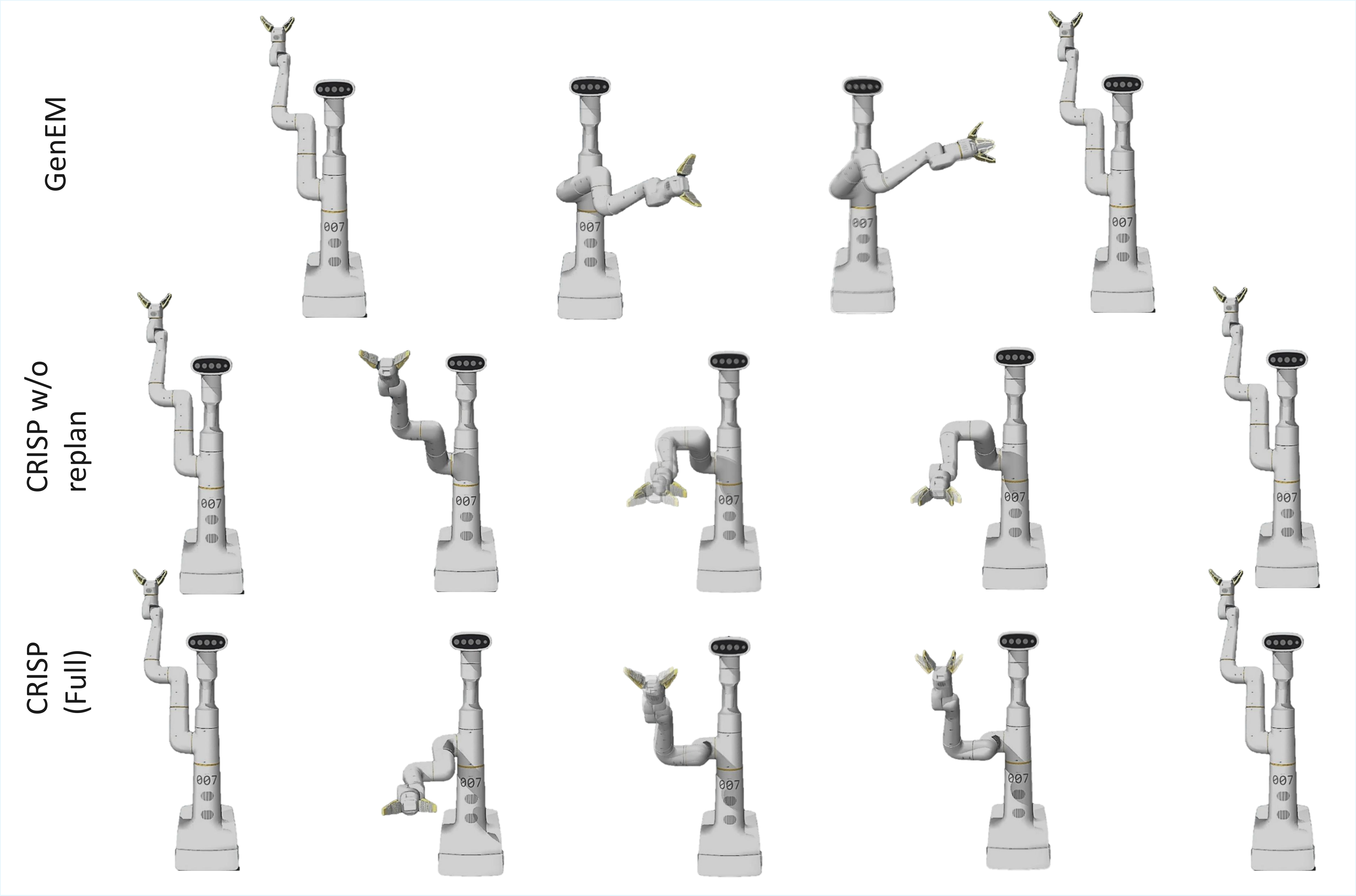}
  {\small \textbf{Scenario:} ``A person is approaching you as a bicycle is about to speed past in front of them.''}\\
  \includegraphics[width=8.5cm]{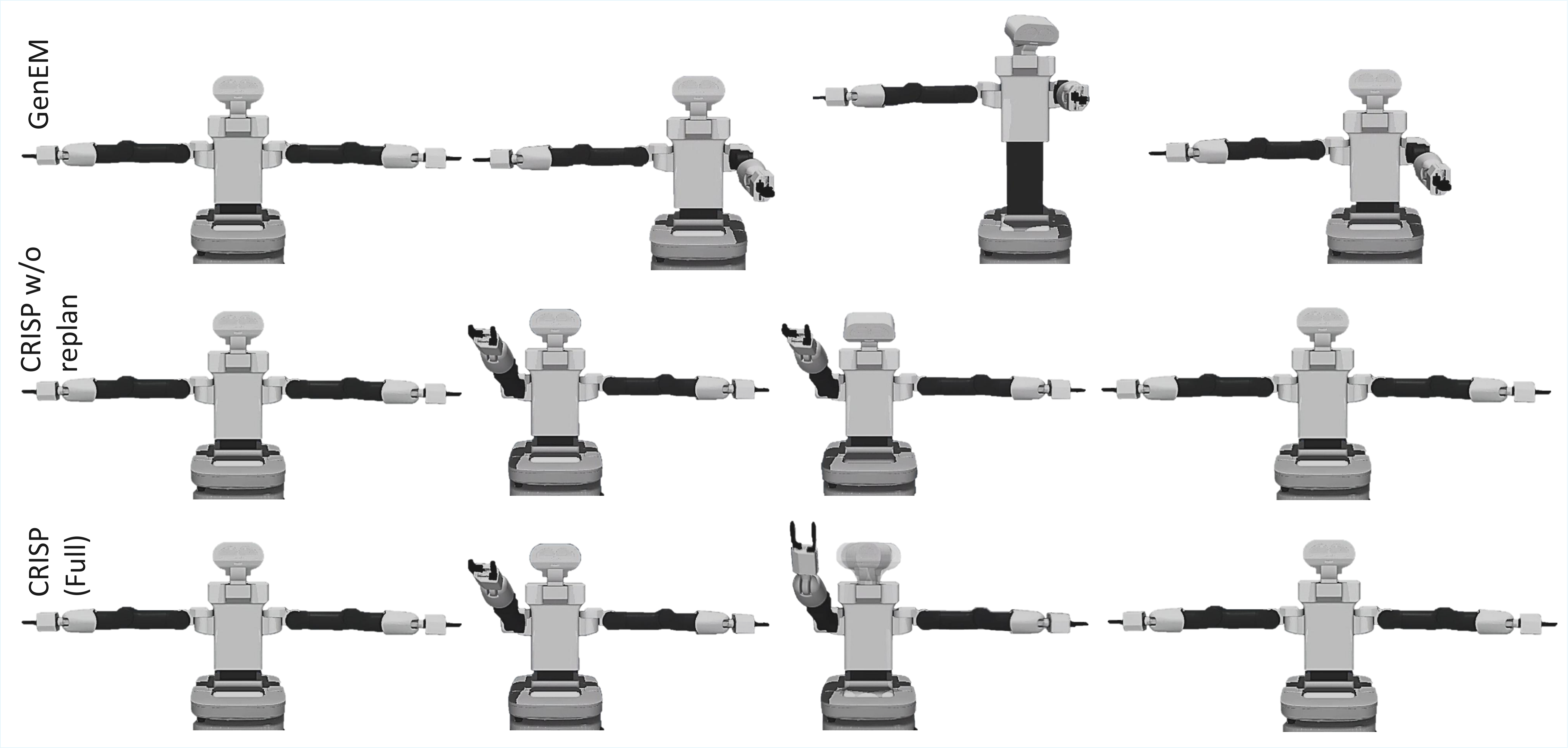}
  \caption{Sample results of generated motions for Everyday robot and TIAGo for given scenarios. The first row shows the result from GenEM, the second row from CRISP w/o replan, and the third row from CRISP (Full). See Supplementary Video.}
  \label{fig:g1result}
  \vspace{-5mm}
\end{figure}

For each robot, we designed four types of scenarios to evaluate behavior generation capabilities from multiple perspectives: 1. Simple, well-defined scenarios (e.g., ``A person waves their hand to greet you.''). 2. Simple, but open-ended scenarios related to emotional responses (e.g., ``A person sighs deeply in disappointment in front of you.''). 3. Scenarios where a person gives a clear command (e.g., ``A person commands you to move your torso up and down and swing your arms.''). 4. Complex, open-ended scenarios involving ambivalent emotions or difficult actions (e.g., ``A person waits for an important online result, fidgeting excitedly with a hopeful expression. After seeing the result on the screen, their face falls, their shoulders slump, and they sink into their chair.''). In these scenarios, we evaluated the performance of our complete system, CRISP with replan, against two baselines. The first baseline was the GenEM model \cite{mahadevan2024generative}, which we modified for a fair comparison. Specifically, instead of using its native Robot API to generate code, we adapted the model to produce direct joint-level commands using the joint information (Joint Set and Joint Limits) provided by our Robot Structure Analyzer. The second baseline was a version of our system without its core components: the Motion Evaluator and the replanning loop (CRISP w/o replan).

To present the evaluation materials to participants, all behaviors corresponding to each robot, scenario, and system combination were recorded as simulation video clips. To ensure consistency, all videos were filmed with the same background and from the same position. After watching each video clip, participants rated the robot's behavior on a 7-point Likert scale across three metrics: (Q1) ease of understanding, (Q2) situational appropriateness, and (Q3) likeability. To prevent fatigue and loss of concentration that could arise from one participant evaluating all five robots, each participant was randomly assigned to evaluate three of the five robots. We used a balanced incomplete block design to ensure fair distribution of robots among participants and a balanced Latin square design to minimize bias from the order of video presentation. Consequently, each participant completed 108 evaluations in total (3 robots $\times$ 4 scenarios $\times$ 3 systems $\times$ 3 questions).

\begin{figure}[t!]
  \centering
  \includegraphics[width=8.5cm]{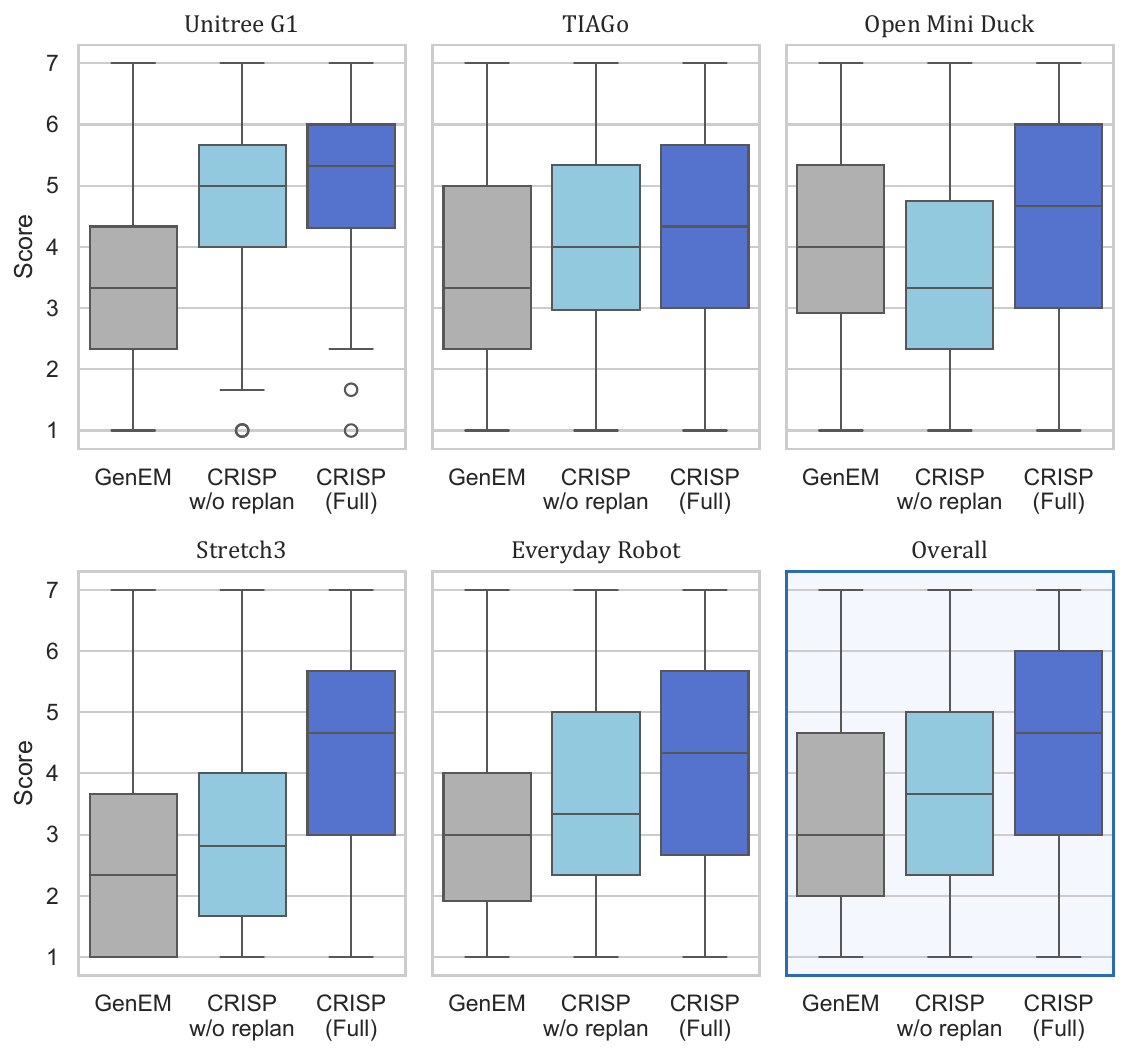}
  \vspace{-2mm}
  \caption{User study results comparing the baseline GenEM, our method without replanning (CRISP without replan), and the full proposed method (CRISP). The box plot labeled `Overall' (bottom right) represents the overall average for each system. Scores are aggregated across Q1, Q2, and Q3.}
  \label{fig:results}
  \vspace{-4mm}
\end{figure}

\subsection{Human Study Results}

We collected responses through the crowdsourcing platform Prolific. The survey was conducted with 50 participants from English-native countries (28 male, 22 female; mean age 42.2 $\pm$ 11.5 years). A simple attention check was included to exclude inattentive participants.
Fig.~\ref{fig:results} presents the mean scores aggregated across all scenarios and metrics (Q1--Q3) for each robot. 
Given the high consistency across Q1--Q3 (Cronbach’s $\alpha > .92$), results are presented as an aggregate score of the three metrics.
The `Overall' plot represents the overall average for each system. The analysis revealed that CRISP had a clear advantage in user preference over the other two systems. The mean preference scores were highest for CRISP at 4.5 ($\pm$1.11), followed by CRISP w/o replan at 3.79 ($\pm$1.16), and GenEM (modified version) at 3.4 ($\pm$1.13). To verify the significance of the difference in preference among the three systems, a Wilcoxon signed-rank test with Holm-Bonferroni correction was performed for pairwise comparisons, which confirmed statistically significant differences between all pairs ($p<.001$). 

Question-wise analysis showed that the mean scores across the three questions were broadly similar (GenEM: Q1=3.31, Q2=3.46, Q3=3.41; CRISP w/o replan: Q1=3.77, Q2=3.84, Q3=3.75; CRISP: Q1=4.50, Q2=4.64, Q3=4.35). Importantly, the same ranking (CRISP $>$ CRISP w/o replan $>$ GenEM) was consistently observed for all three questions, underscoring the robustness of the improvement.

\begin{figure}[t!]
  \centering
  {\small \textbf{Scenario:} ``A person is looking for a seat, and you guide them to an empty one on your right.''} \\
  \includegraphics[width=8.5cm]{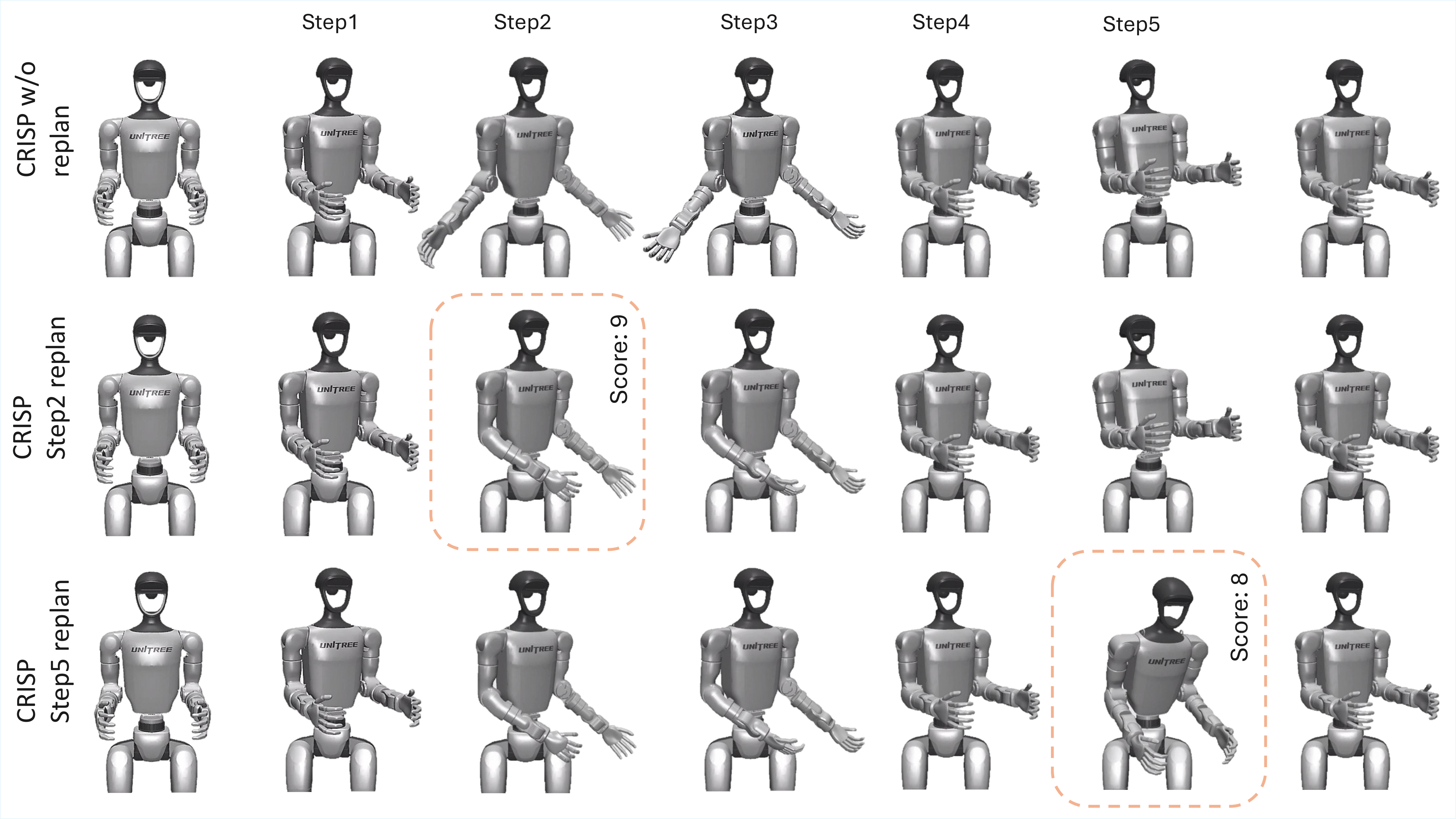}
  \vspace{-3mm}
  \caption{Evolution of a plan through replanning. The first row is the initial plan (CRISP w/o replan). The second and third rows show subsequent revisions. The boxed steps are the actions modified during replanning (at steps 2 and 5), with the score in each box indicating the reward assigned by the VLM for the new action.}
  \label{fig:replan}
  \vspace{-3mm}
\end{figure}

Please see Fig. \ref{fig:g1result} and supplementary video for the sample results. Fig. \ref{fig:replan} shows how the behavior evolves through replanning.

\subsection{Ablation Study}
\label{ssec:ab}
\begin{figure}[!t]
  \centering
  \includegraphics[width=8.5cm]{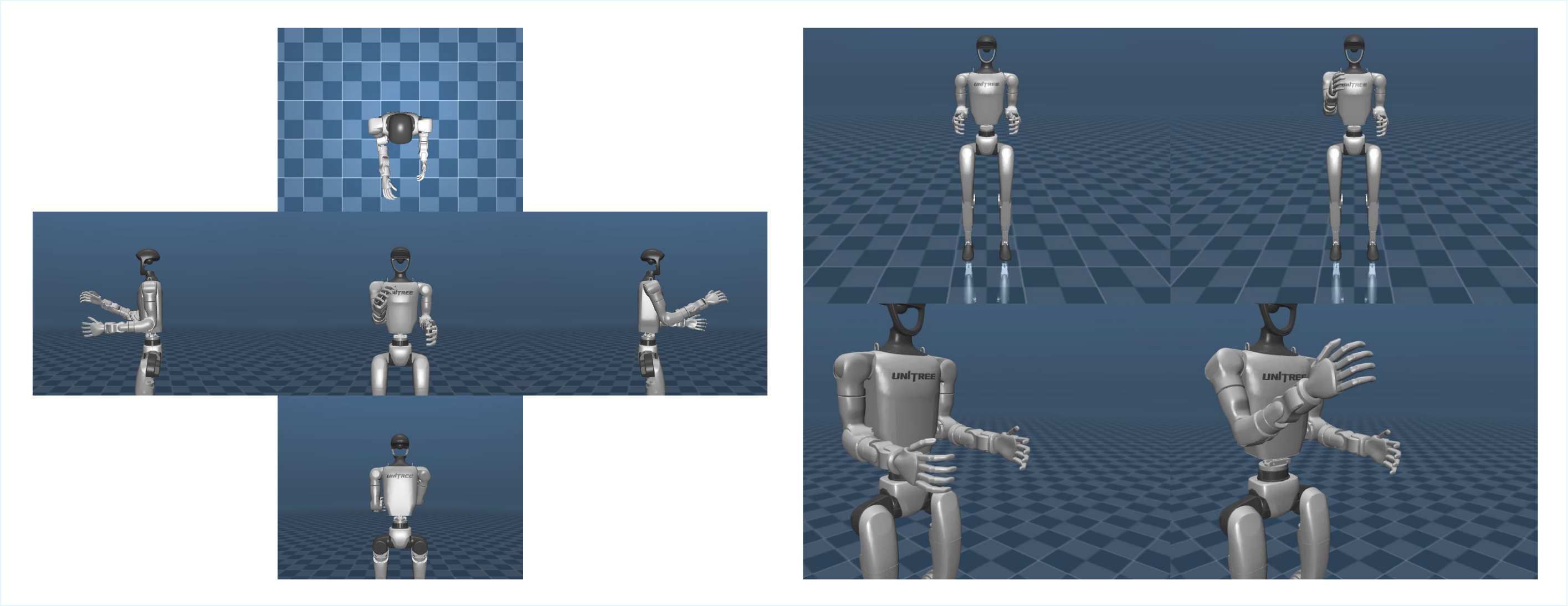}
  \vspace{-3mm}
  \caption{Example VLM input images. (Left) Temporally-sampled multi-view input for the M3 model. (Right) Our proposed keyframe-based, dual-view (full and zoom shot) input.}
  \label{fig:multiview}
  \vspace{-4mm}
\end{figure}

To verify the impact of each component of our proposed system on its overall performance, we conducted an ablation study. We compared CRISP with four variant models to evaluate the contribution and necessity of each component: (M1) a model where the Behavior Refiner generates only a single action, (M2) a model that refines the entire plan at once, (M3) a model that takes multi-view images (Fig. \ref{fig:multiview}) as input, and (M4) a model that excludes the zoomed-in shot from the standard VLM input (Fig. \ref{fig:keyframe}). For a fair comparison, the number of replan iterations for all models was limited to 10, and the same Unitree G1 robot was used. Ten evaluators watched five generated videos per model for the scenario ``A person waves their hand to greet you'' and rated the task success as `Success', `Unsure', or `Failure'. For analysis, `Unsure' was treated as `Failure'. The success count in Table \ref{tab:ablation} is calculated by summing all `Success' ratings from the ten evaluators and dividing by the number of runs (five).

\begin{table}[h!]
\centering
\caption{Ablation study results. Average values over 5 runs.}
\label{tab:ablation}
\begin{tabular}{l c c c}
\toprule
\textbf{Model} & \textbf{Success Count} & \textbf{\# VLM Input Images} & \textbf{\# Replans} \\
\midrule
Proposed  & 4.6 & 10.2 & 3.4 \\
M1        & 1.7 & 9.4  & 9.4 \\
M2        & 0.9 & 140.0 & 10.0 \\
M3        & 3.7 & 41.2 & 3.4 \\
M4        & 2.6 & 13.8 & 4.4 \\
\bottomrule
\end{tabular}
\end{table}

Table \ref{tab:ablation} shows the results of the ablation study. The proposed system achieved an average success count of 4.6 out of 5 runs, using an average of 10.2 VLM input images and 3.4 replan iterations to achieve its goal, demonstrating high efficiency.
The utility of each module in the proposed system can be confirmed by comparing it with the variant models. M1, which only refines a single action, had a similar number of input images (9.4) to the proposed model but required 9.4 replan iterations and achieved only 1.7 successes. This suggests it fell into a local optimum and performed inefficient searches. M2, which refined the entire plan at once, had the lowest success count of 0.9, used 140.0 input images, and reached the maximum of 10.0 replan iterations, demonstrating its significant inefficiency. M3, using multi-view images, showed a relatively high success rate of 3.7 but was inefficient, requiring 41.2 images to achieve its goal--more than four times the visual information needed by the proposed model. Finally, M4, which excluded the crucial zoomed-in shots, saw its success count drop to 2.6, clearly showing that accurately perceiving the robot's fine movements is vital for task success.

\section{Discussion}
The proposed replan model demonstrated its superiority by achieving the highest user preference in most scenarios. CRISP's performance improvement was especially prominent with the physically complex Unitree G1, suggesting that our use of visual information plays a key role in planning subtle postures and movements for complex robots.

Despite the overall superiority of the replan model (average user rating CRISP: 4.5, GenEM: 3.4), exceptional results were observed in scenario 2 for the Open Mini Duck (CRISP: 3.9, GenEM: 4.4) and TIAGo robots (Open Mini Duck: ``A person is in front of you with a sad expression.'', TIAGo: ``A person cowers in fear in front of you.''). These results are likely attributable to the unique nature of scenario 2, which involves open-ended social interactions. This scenario evaluates the robot's response to user emotions, and human expectations for a response to emotions like sadness are subjective and varied. For example, some users might expect empathy and comfort, while others might prefer a humorous action to change the mood. In such situations where there is no single correct answer, it is difficult for a VLM-generated behavior plan to satisfy every user's individual expectations. This contrasts with the universally high performance of the models in scenario 3, which had a clear objective. This indicates that while the proposed system is highly effective for tasks with objective goals, adapting to the subjective and complex social contexts of humans is a challenge that needs to be addressed in future research.

Additionally, several technical aspects suggest directions for future work. First, as current VLMs evaluate spatial accuracy via static images, they lack temporal awareness (e.g., speed or smoothness). Future research should incorporate video processing to refine these temporal elements. Second, while this study focused on upper-body interaction, whole-body dynamics like walking remain beyond the current system's scope. Implementing such behaviors would likely require a hybrid approach that integrates our framework with physics-informed control or existing API-based methods.

Regarding implementation, the current system takes $\sim$3 min for initial motion generation (Joint Structure Analyzer: $\sim$1 min, Robot Behavior Generator: $\sim$20 s, Joint Code Generator: $\sim$1.5 min). The replanning process requires 10--20 min depending on the complexity and iterations (avg. $\sim$2.5 min per replan), making it ideally suited for offline robot behavior authoring. This allows developers to pre-generate high-quality social motions without manual key-framing. Moreover, the framework establishes an autonomous pipeline for creating large-scale training datasets, which can be applied to Vision-Language-Action (VLA) models to facilitate real-time generation. Ultimately, this research serves as a vital foundation for securing the real-time social interaction capabilities of robots, advancing VLM-based behavior generation toward practical, real-world deployment.

\section{Conclusion}
In this paper, we presented CRISP, an autonomous replanning framework that employs a VLM as a social critic to enable robotic self-correction of social behaviors. By leveraging low-level control based on robot structure files, the framework achieves high flexibility and generalizability. Our user studies with five distinct robot types validated that CRISP generates more appropriate and socially preferred behaviors than existing methods, and an ablation study confirmed the efficacy of each component. 
We believe the proposed framework marks a step forward toward socially aware robots that autonomously adapt and refine their behaviors in human-centric environments, reducing reliance on costly human feedback and broadening cross-platform applicability.
Remaining challenges include handling subjective social contexts, improving temporal expressiveness, and achieving real-time performance. Addressing these limitations will further advance the development of interactive robots capable of rich, autonomous social presence.



\section*{Acknowledgments}
Illustrations in Fig. 1 were generated by Google Gemini.

\bibliographystyle{IEEEtran}
\bibliography{references}

@inproceedings{mohammadi2019designing,
  title={Designing a personality-driven robot for a human-robot interaction scenario},
  author={Mohammadi, Hadi Beik and Xirakia, Nikoletta and Abawi, Fares and Barykina, Irina and Chandran, Krishnan and Nair, Gitanjali and Nguyen, Cuong and Speck, Daniel and Alpay, Tayfun and Griffiths, Sascha and others},
  booktitle={ICRA},
  year={2019},
  organization={IEEE}
}

@inproceedings{porfirio2020transforming,
  title={Transforming robot programs based on social context},
  author={Porfirio, David and Saupp{\'e}, Allison and Albarghouthi, Aws and Mutlu, Bilge},
  booktitle={CHI},
  year={2020}
}

@inproceedings{mahadevan2024generative,
  title={Generative expressive robot behaviors using large language models},
  author={Mahadevan, Karthik and Chien, Jonathan and Brown, Noah and Xu, Zhuo and Parada, Carolina and Xia, Fei and Zeng, Andy and Takayama, Leila and Sadigh, Dorsa},
  booktitle={HRI},
  year={2024}
}

@article{park2024towards,
  title={Towards Embedding Dynamic Personas in Interactive Robots: Masquerading Animated Social Kinematic (MASK)},
  author={Park, Jeongeun and Jeong, Taemoon and Kim, Hyeonseong and Byun, Taehyun and Shin, Seungyoun and Choi, Keunjun and Kwon, Jaewoon and Lee, Taeyoon and Pan, Matthew and Choi, Sungjoon},
  journal={IEEE Robotics and Automation Letters},
  year={2024},
  publisher={IEEE}
}

@article{destephe2015walking,
  title={Walking in the uncanny valley: Importance of the attractiveness on the acceptance of a robot as a working partner},
  author={Destephe, Matthieu and Brandao, Martim and Kishi, Tatsuhiro and Zecca, Massimiliano and Hashimoto, Kenji and Takanishi, Atsuo},
  journal={Frontiers in psychology},
  volume={6},
  pages={204},
  year={2015},
  publisher={Frontiers Media SA}
}

@inproceedings{shentu2024llms,
  title={From llms to actions: Latent codes as bridges in hierarchical robot control},
  author={Shentu, Yide and Wu, Philipp and Rajeswaran, Aravind and Abbeel, Pieter},
  booktitle={IROS},
  year={2024},
}

@inproceedings{williams2017information,
  title={Information theoretic MPC for model-based reinforcement learning},
  author={Williams, Grady and Wagener, Nolan and Goldfain, Brian and Drews, Paul and Rehg, James M and Boots, Byron and Theodorou, Evangelos A},
  booktitle={ICRA},
  year={2017}
}

@InProceedings{kim2024openvla,
  title = 	 {OpenVLA: An Open-Source Vision-Language-Action Model},
  author =       {Kim, Moo Jin and Pertsch, Karl and Karamcheti, Siddharth and Xiao, Ted and Balakrishna, Ashwin and Nair, Suraj and Rafailov, Rafael and Foster, Ethan P and Sanketi, Pannag R and Vuong, Quan and Kollar, Thomas and Burchfiel, Benjamin and Tedrake, Russ and Sadigh, Dorsa and Levine, Sergey and Liang, Percy and Finn, Chelsea},
  booktitle = 	 {CoRL},
  year = 	 {2025},
}

@article{mei2024replanvlm,
  title={Replanvlm: Replanning robotic tasks with visual language models},
  author={Mei, Aoran and Zhu, Guo-Niu and Zhang, Huaxiang and Gan, Zhongxue},
  journal={IEEE Robotics and Automation Letters},
  year={2024},
  publisher={IEEE}
}

@article{chen2025robogpt,
  title={Robogpt: an llm-based long-term decision-making embodied agent for instruction following tasks},
  author={Chen, Yaran and Cui, Wenbo and Chen, Yuanwen and Tan, Mining and Zhang, Xinyao and Liu, Jinrui and Li, Haoran and Zhao, Dongbin and Wang, He},
  journal={IEEE Transactions on Cognitive and Developmental Systems},
  year={2025},
  publisher={IEEE}
}

@inproceedings{dong2024survey,
  title={A survey on in-context learning},
  author = "Dong, Qingxiu  and
      Li, Lei  and
      Dai, Damai  and
      Zheng, Ce  and
      Ma, Jingyuan  and
      Li, Rui  and
      Xia, Heming  and
      Xu, Jingjing  and
      Wu, Zhiyong  and
      Chang, Baobao  and
      Sun, Xu  and
      Li, Lei  and
      Sui, Zhifang",
  booktitle={EMNLP},
  year={2024}
}

@article{skreta2024replan,
  title={Replan: Robotic replanning with perception and language models},
  author={Skreta, Marta and Zhou, Zihan and Yuan, Jia Lin and Darvish, Kourosh and Aspuru-Guzik, Al{\'a}n and Garg, Animesh},
  journal={arXiv:2401.04157},
  year={2024}
}

@inproceedings{suguitan2020moveae,
  title={MoveAE: modifying affective robot movements using classifying variational autoencoders},
  author={Suguitan, Michael and Gomez, Randy and Hoffman, Guy},
  booktitle={HRI},
  year={2020}
}

@article{oralbayeva2024data,
  title={Data-driven communicative behaviour generation: A survey},
  author={Oralbayeva, Nurziya and Aly, Amir and Sandygulova, Anara and Belpaeme, Tony},
  journal={ACM Transactions on Human-Robot Interaction},
  volume={13},
  number={1},
  pages={1--39},
  year={2024},
  publisher={ACM New York, NY}
}

@inproceedings{aly2013model,
  title={A model for synthesizing a combined verbal and nonverbal behavior based on personality traits in human-robot interaction},
  author={Aly, Amir and Tapus, Adriana},
  booktitle={HRI},
  year={2013},
  organization={IEEE}
}

@inproceedings{huang2012robot,
  title={Robot behavior toolkit: generating effective social behaviors for robots},
  author={Huang, Chien-Ming and Mutlu, Bilge},
  booktitle={HRI},
  year={2012}
}

@inproceedings{david2022interaction,
  title={Interaction templates: a data-driven approach for authoring robot programs},
  author={David, Porfirio and Cakmak, Maya and Saupp{\'e}, Allison and Albarghouthi, Aws and Mutlu, Bilge},
  booktitle={PLATEAU: 12th Annual Workshop at theIntersection of PL and HCI},
  year={2022}
}

@article{tian2024maximizing,
  title={Maximizing alignment with minimal feedback: Efficiently learning rewards for visuomotor robot policy alignment},
  author={Tian, Ran and Wu, Yilin and Xu, Chenfeng and Tomizuka, Masayoshi and Malik, Jitendra and Bajcsy, Andrea},
  journal={arXiv:2412.04835},
  year={2024}
}

@article{huang2025emotion,
  title={Emotion: Expressive motion sequence generation for humanoid robots with in-context learning},
  author={Huang, Peide and Hu, Yuhan and Nechyporenko, Nataliya and Kim, Daehwa and Talbott, Walter and Zhang, Jian},
  journal={IEEE Robotics and Automation Letters},
  year={2025},
  publisher={IEEE}
}

@inproceedings{wei2022chain,
  title={Chain-of-thought prompting elicits reasoning in large language models},
  author={Wei, Jason and Wang, Xuezhi and Schuurmans, Dale and Bosma, Maarten and Xia, Fei and Chi, Ed and Le, Quoc V and Zhou, Denny and others},
  booktitle={NIPS},
  year={2022}
}

@inproceedings{weber2018shape,
  title={How to shape the humor of a robot-social behavior adaptation based on reinforcement learning},
  author={Weber, Klaus and Ritschel, Hannes and Aslan, Ilhan and Lingenfelser, Florian and Andr{\'e}, Elisabeth},
  booktitle={ICMI},
  year={2018}
}

@article{wang2025multi,
  title={Multi-Agent LLM Actor-Critic Framework for Social Robot Navigation},
  author={Wang, Weizheng and Obi, Ike and Min, Byung-Cheol},
  journal={arXiv:2503.09758},
  year={2025}
}

@INPROCEEDINGS{yoon2019robots,
  author={Yoon, Youngwoo and Ko, Woo-Ri and Jang, Minsu and Lee, Jaeyeon and Kim, Jaehong and Lee, Geehyuk},
  booktitle={ICRA}, 
  title={Robots Learn Social Skills: End-to-End Learning of Co-Speech Gesture Generation for Humanoid Robots}, 
  year={2019}
}

\end{document}